\documentclass{article}

\usepackage{arxiv}

\usepackage[utf8]{inputenc} 
\usepackage[T1]{fontenc}    
\usepackage{hyperref}       
\usepackage{url}            
\usepackage{booktabs}       
\usepackage{amsfonts}       
\usepackage{nicefrac}       
\usepackage{microtype}      
\usepackage{lipsum}
\usepackage{graphicx}
\graphicspath{ {./images/} }

\usepackage{amsmath}
\usepackage{graphicx}
\usepackage{caption}
\usepackage{subcaption}
\usepackage{csquotes}
\graphicspath{{./figures/}}
\everymath{\displaystyle}
\usepackage{comment}
\usepackage{amssymb} 
\usepackage[table,xcdraw]{xcolor}
\usepackage{booktabs}
\usepackage{graphicx}
\usepackage{varwidth}
\usepackage{rotating}
\usepackage{multicol}

\usepackage[table]{xcolor} 
\usepackage{colortbl}      
\usepackage{hyperref}
\usepackage{csquotes}
\MakeOuterQuote{"}
\usepackage{placeins}
\usepackage{makecell} 
\usepackage{array} 
\usepackage{cite}

\title{RECALL-MM: A Multimodal Dataset of Consumer Product Recalls for Risk Analysis using Computational Methods and Large Language Models}

\author{
Diana Bolanos \\
  Department of Mechanical Engineering\\
	University of California\\
	Berkeley, California 94720\\
  \texttt{dbolanos@berkeley.edu} \\
   \And
Mohammadmehdi Ataei \\
Autodesk Research\\
Toronto, ON M5G 1M1, Canada\\
  \texttt{mehdi.ataei@autodesk.com} \\
  \And
Daniele Grandi \\
  Autodesk Research\\
  San Francisco, CA 94105\\
  \texttt{daniele.grandi@autodesk.com} \\
\And
Kosa Goucher-Lambert\\
Department of Mechanical Engineering\\
	University of California\\
	Berkeley, California 94720\\
\texttt{kosa@berkeley.edu} \\
}

\begin{document}
\maketitle
\begin{abstract}
Product recalls provide valuable insights into potential risks and hazards within the engineering design process, yet their full potential remains underutilized. In this study, we curate data from the United States Consumer Product Safety Commission (CPSC) recalls database to develop a multimodal dataset, RECALL-MM, that informs data-driven risk assessment using historical information, and augment it using generative methods. Patterns in the dataset highlight specific areas where improved safety measures could have significant impact. We extend our analysis by demonstrating interactive clustering maps that embed all recalls into a shared latent space based on recall descriptions and product names. Leveraging these data-driven tools, we explore three case studies to demonstrate the dataset's utility in identifying product risks and guiding safer design decisions. The first two case studies illustrate how designers can visualize patterns across recalled products and situate new product ideas within the broader recall landscape to proactively anticipate hazards. In the third case study, we extend our approach by employing a large language model (LLM) to predict potential hazards based solely on product images. This demonstrates the model's ability to leverage visual context to identify risk factors, revealing strong alignment with historical recall data across many hazard categories. However, the analysis also highlights areas where hazard prediction remains challenging, underscoring the importance of risk awareness throughout the design process. Collectively, this work aims to bridge the gap between historical recall data and future product safety, presenting a scalable, data-driven approach to safer engineering design. 
\end{abstract}

\section{Introduction}
Risk analysis is a necessary step in the product development process. Engineers and designers are encouraged to predict potential hazards using traditional six-sigma approaches to assess potential failure modes \cite{ulrich2016product}. Nonetheless, consumer products are often recalled due to design and manufacturing related hazards, posing a risk of injury and sometimes death~\cite{hora2011safety}. As such, we see an opportunity to learn from recalled products to observe \textit{what} products fail, and \textit{how} the failures occur, ultimately providing engineers and designers historical information of existing failure modes.
This study leverages the United States Consumer Product Safety Commission (CPSC) recalls database to serve as a benchmark for novel computational risk prediction approaches presented herein. We curate a dataset of 6,874 recalls spanning dates between the years 2000 and 2024, augmenting the retrieved database information with new descriptors created using a large language model (LLM).  Notably, these recalls account for over 546 million individual product SKUs reported as affected over the past two decades, underscoring the vast scale and real-world impact of product safety failure. An example of preprocessed recall entry can be found in Appendix A Table \ref{tab:recall_data}. We highlight the use of the dataset and present how it could support risk identification in the design process.

Our contributions include: (1) the development of RECALL-MM, a curated, multimodal dataset of recalled consumer products, augmented through LLM-generated classifications and visual descriptors, (2) the demonstration of computational methods for embedding and visualizing recall data to uncover patterns in product failures, supported by two case studies illustrating how these methods can aid risk identification, and (3) the application of an LLM to predict potential product hazards based solely on visual descriptions, highlighting both the strengths and limitations of automated hazard assessment. Collectively, this work aims to improve product safety, anticipate design failures, and support data-driven decision-making in engineering design.

To support further development and evaluation of our dataset, we make the RECALL-MM dataset and accompanying experimental code publicly available on GitHub\footnote{\href{https://github.com/dianabolanos/RECALL-MM}{https://github.com/dianabolanos/RECALL-MM}}.

\section{Related Work}
\label{sec:headings}
Over the last decade, several large design datasets have been curated and released to support data-driven design efforts for product design and other design tasks. The classes of objects collected, sample size, and modality of the data are the main differentiators between datasets in this field.

Shapenet~\cite{shapenet2015}, the ABC Dataset~\cite{Koch_2019_CVPR}, DeepCAD~\cite{Wu_Xiao_Zheng_2021}, and the Fusion 360 Gallery dataset~\cite{willis2021joinable} are among the largest datasets that contain geometry data and class labels for individual parts and whole assemblies. The datasets have been widely used for design automation and geometry generation tasks, and have also supported other work around ancillary design tasks such as materials selection~\cite{bian2024hg}. 
Other smaller datasets have also been curated around more specific classes of objects, such as car bodies, mechanical components, and bicycles~\cite{kim2020large, regenwetter2024biked++,NEURIPS2024_013cf29a}. 
While these datasets provide valuable structured information on product features, making them a useful starting point for analysis, they also come with limitations, including incorrect or missing semantic information, limited number of object classes, and lack of design context, intent, and criteria, which limits their utility for comprehensive risk assessment.

Other design datasets have focused on different modalities, such as hand-drawn sketches~\cite{Toh_Miller_2016, Gryaditskaya_Sypesteyn_Hoftijzer_Pont_Durand_Bousseau_2019}, or textual descriptions of designs~\cite{Zhu_Luo_2022, Meltzer_Lambourne_Grandi_2023, Jiang_Sarica_Song_Hu_Luo_2022, Goucher-Lambert_Cagan_2019}. Although these datasets focus more on design rationale, describing the features and aesthetics of the design solutions, they do not explicitly consider design feasibility, and some are limited by the number of object classes and data quality.

A few multimodal design datasets have been published, built around graphic design~\cite{Lin_Huang_Zhao_Zhan_Lin_2024}, design requirement documents~\cite{Doris_Grandi_Tomich_Alam_Ataei_Cheong_Ahmed_2024}, and descriptions of design changes~\cite{achlioptas2023shapetalk}. These datasets combine textual descriptions, geometry, images, and design requirements in various combinations to support design tasks such as editing 3D geometry, interpreting design requirements, and style classification.

The multimodal dataset collected in this work differs from prior datasets as it provides textual descriptions of designs, images, and recall information related to a failure mode of the product over a wide range of product classes.

\subsection{Data-Driven Risk Analysis}

Product recalls have been used to investigate trends in consumer product safety, with a prevailing focus on children's toys \cite{durrett2015decade, niven2020hazardous, kirschman2007resale, wai2024effectiveness}. These studies similarly leverage the CPSC database, along with the global recalls dataset from the Organisation for Economic Co-operation and Development (OECD) \cite{oecd_global_recalls}. Wai and Uttama \cite{wai2024effectiveness} present a series of machine learning approaches for predicting a binary classification of children's toy safety, showcasing the potential for data-driven hazard prediction. Our study expands on this work by investigating trends across multiple product categories, moving beyond a single domain focus.

The automotive industry has also seen a shift towards data-driven design for predictive maintenance and hazard prevention. Yorulmus et al. \cite{yorulmucs2022predictive} present machine learning approaches for predicting brake defects from vehicles passing quality assurance checkpoints, yet exhibiting high rates of customer complaints. Similarly, \cite{o2024determination} demonstrated a multi-label transfer learning approach by implementing a series of pretrained Convolutional Neural Networks (CNNs) to predict the binary failure status of multiple engine components. Both studies rely on failure data to improve future design of automotive components, demonstrating the effectiveness of leveraging data for failure mitigation and design improvement.

\subsection{Hazard Identification in Design}
The field of risk analysis has been extensively studied and continues to evolve within various organizational and engineering domains.  A fundamental objective of risk analysis in engineering design is to proactively identify and mitigate hazards before they manifest as failures or safety incidents. Failure Mode and Effects Analysis (FMEA) is among the most widely adopted methodologies employed by organizations to structure systematic risk assessments, prioritize potential hazards, and implement preventive actions \cite{stamatis2003failure, carlson2012effective, modarres2006risk}. First developed in the 1960s by the aerospace industry \cite{bowles1995fuzzy}, FMEAs now serve as an industry standard tool across various applications, including automotive design, aerospace engineering, and product development, demonstrating its versatility in supporting product reliability and safety. More recent advancements in FMEA methodologies incorporate computational approaches, such as fuzzy logic, machine learning, and integrated decision-making frameworks, enhancing traditional FMEA practices by addressing uncertainties and subjectivity inherent in risk scoring \cite{liu2013risk}. These improvements continue to reinforce the importance of hazard identification and risk mitigation in complex engineering designs. We further posit that reviewing historical data can expand the results of risk analysis activities.

\section{Methods}
\label{sec:others}
We focus this section on detailing the steps used to clean and augment the CPSC database into a multimodal dataset used throughout the study. Then, we describe the computational methods used to embed this dataset into a vectorized representation, allowing for deeper analysis and visualization. Finally, we introduce the methodology for leveraging an LLM to predict hazards based on visual product information.

\begin{figure}[t]
\begin{center}
\includegraphics[width=0.75\linewidth]{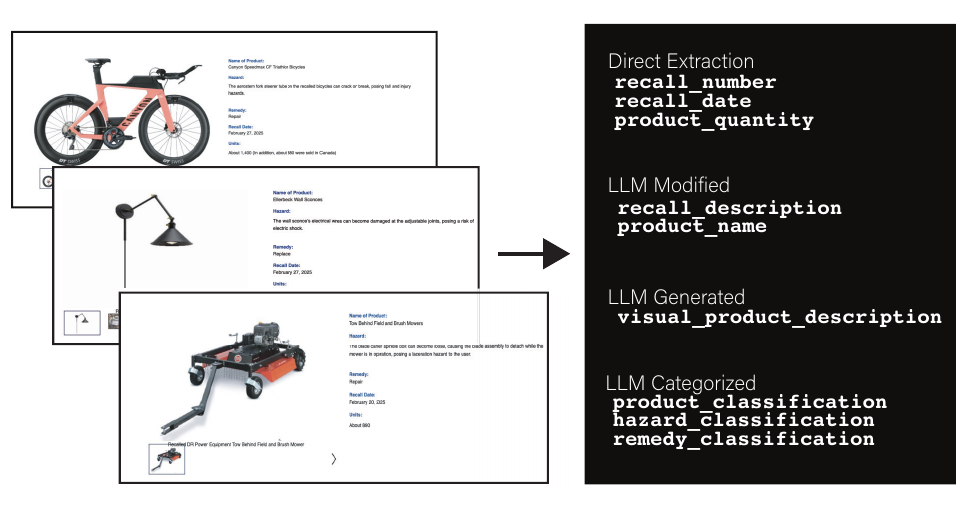}
\caption{Process overview of translating database information into nine distinct data fields. }
\label{fig:overview}
\end{center}
\end{figure}

\subsection{Dataset}
The dataset used in this analysis is a curated subset of the US CPSC recalls database \cite{CPSC_Recalls}. To ensure consistency and feasibility for analysis, recalls were filtered based on API accessibility, the presence of product images, and the timeframe of 2000 to 2024. This filtering process resulted in a dataset comprising 6,874 recall entries. While this represents only a portion of all CPSC recalls, it remains sufficiently comprehensive and reflects broader trends observed in publicly available aggregate data \cite{CPSC_Recalls, oecd_global_recalls}.

Each entry in the dataset includes essential recall attributes such as hazard classifications, product categories, remedy types, historical recall dates, and an associated product image. The raw CPSC data lacked labeled classifications for each entry. To address this, we leveraged generative models, specifically GPT-4o \cite{openai2024gpt4o}, to enrich and structure the data. Figure \ref{fig:overview} illustrates which fields were directly extracted from the CPSC database and which were augmented using an LLM.

The  \textbf{LLM Modified} fields--\textit{recall\_description} and \textit{product\_name}--were refined using GPT-4o to remove brand references and retain generic descriptions. Additionally, we label the \textit{visual\_product\_description} field as \textbf{LLM Generated}, as it was generated entirely by prompting GPT-4o to describe the product based on its associated image. For classification tasks, GPT-4o assigned each recall to a product category, hazard category, and remedy type, selecting from predefined lists. We label this as \textbf{LLM Categorized}. Hazard and remedy categories were aligned with CPSC’s own labels, with definitions provided in Appendix B (Tables \ref{tab:hazard_classifications} and \ref{tab:remedies_classifications}). Since the CPSC does not offer standardized product categories, we developed an 11-category scheme based on domain understanding, ensuring it broadly covers the landscape of recalled products while maintaining consistency with OECD terminology \cite{oecd_global_recalls}.

During data cleaning, all records were standardized to a predefined schema, ensuring consistent representation of attributes such as product description, hazard type, and remedy details. GPT-4o outputs were validated against this schema, with type and value constraints applied to ensure reliability. Each record retains its original recall ID, preserving traceability to the source data and supporting future reference or verification.

\subsection{Recall Space Exploration and Visualization}

\subsubsection{Embedding}
Product descriptors, specifically \textit{recall\_description} and \textit{product\_name}, were embedded into a numerical latent space using the \texttt{all-MiniLM-L6-v2} model from Sentence-BERT \cite{reimers2019sentence}. Sentence-BERT is a pre-trained model that generates fixed-length dense vector representations optimized for capturing semantic similarity between text inputs. We selected the \texttt{all-MiniLM-L6-v2} model as it offers an effective balance between model size, computational efficiency, and embedding quality, making it well-suited for large-scale analyses without sacrificing performance.

\subsubsection{Dimensionality Reduction for Visualization}
To visualize relationships among products and recall reasons, we applied dimensionality reduction techniques. Specifically, we employed the t-distributed Stochastic Neighbor Embedding (t-SNE) algorithm, chosen for its strength in preserving local structure and effectively capturing complex, non-linear relationships in high-dimensional data \cite{van2008visualizing}. Compared to linear methods such as Principal Component Analysis (PCA) \cite{abdi2010principal}, which primarily maintain global variance, t-SNE excels at revealing dense clusters and neighborhood groupings, which are paramount features for identifying semantically similar products and localized recall patterns. Product embeddings obtained via Sentence-BERT were projected from their original vector space into two and three-dimensional coordinates. The resulting visual maps (Fig. \ref{fig:example_2d} and Fig. \ref{fig:3d_embeddings}) enable exploration of recall clusters, offering insight into product similarities and hazard trends. While other methods like UMAP \cite{mcinnes2018umap} could also be considered, we prioritized t-SNE for its well-established use in exploratory visualizations where fine-grained local structure is of primary interest.

\subsection{ LLM-Based Hazard Prediction}
In addition to computational exploration of the recall data, we evaluate the feasibility of using LLMs to predict potential hazards directly from product images. Specifically, we focus on the \textit{visual\_product\_description} field, which contains a textual description of each product image generated by GPT-4o.

To perform hazard prediction, we prompt an LLM to analyze the textual description of the image and output all applicable hazard classifications. The model selects hazard labels from a predefined set of ten hazard categories (Appendix B, Table \ref{tab:hazard_classifications}), ensuring consistency with existing CPSC classifications. The full prompt used for LLM prediction is provided: 

\begin{quote}
\noindent\texttt{You are a product safety expert. Identify all potential hazards for the given product. Provide output in valid JSON format only, structured as:}

\medskip

\texttt{\{ \\
\hspace*{1em} `product': \{product\_description\},\\
\hspace*{1em} `predicted\_hazards': [List of all applicable hazards]\\
\}}

\medskip

\noindent\texttt{The predicted\_hazards field must only contain hazards from this set: \{all\_hazards\}. Do not leave any predicted\_hazards fields empty. If multiple hazards apply, include all relevant ones. No explanations - return JSON only.}
\end{quote}

Outputs are returned in a strict JSON format, listing the product description and the predicted hazards. The prompt enforces that the model must not leave any hazard fields empty, encouraging comprehensive identification of potential risks.

\subsubsection*{Evaluation Metric}
To quantify the model's performance, we introduce a Relaxed Accuracy (RA) metric:

\begin{equation}
\delta_i = 
\begin{cases}
1, & \text{if } g_i \in P_i \\
0, & \text{otherwise}
\end{cases}
\end{equation}

\begin{equation}
\text{Relaxed Accuracy (RA)} = \frac{1}{N} \sum_{i=1}^{N} \delta_i
\label{RA}
\end{equation}

where \\
\noindent
\begin{itemize}
    \item $N$ = Total number of products per hazard class.
    \item $g_i$ = Single ground truth hazard classification for product $i$.
    \item $P_i$ = Set of predicted hazard classifications for product $i$.
    \item $\delta_i$ = Indicator function that equals 1 if the ground truth hazard $g_i$ is present in the predicted set $P_i$, and 0 otherwise.
\end{itemize}

The RA metric accounts for cases where the LLM predicts multiple hazards, but the recall dataset provides only one ground truth hazard per entry. As such, the metric does not penalize overprediction, reflecting the real-world possibility of multiple concurrent hazards.

\section{Results and Discussion}
To evaluate the reliability of the curated and augmented dataset, we begin by validating the LLM-generated categorizations against human-annotated ground truths, ensuring consistency across product, hazard, and remedy classifications. Following this, we analyze aggregate trends within the dataset,  examining prevalent hazards, product categories, and remedy actions over time. Building on these observations, we then present three case studies to illustrate different approaches for leveraging the dataset in risk identification and design decision-making. The first two case studies employ computational methods, embedding recall data into a shared latent space to explore product relationships and potential risks. The third case study investigates the feasibility of using LLMs to predict potential hazards based solely on visual context.

\subsection{Human Evaluation of Dataset Categorizations}
To validate the reliability of the LLM categorizations, we compare them against ground truth labels derived from three independent human annotators. Each annotator performed 100 classifications for each task, amounting to 900 annotations. To establish a ground truth, we employed majority voting across the three annotators' labels for each classification task (product, hazard, and remedies). We assessed inter-rater reliability by evaluating Fleiss's Kappa, which yielded coefficients of 0.71 for product classification, 0.80 for hazard classification, and 0.85 for remedies classification. These scores indicate substantial to almost perfect agreement, based on standard interpretation thresholds. Given this high level of consistency, majority voting was deemed appropriate to consolidate the annotations. Items where no majority agreement was reached (5 for product, 4 for hazard, and 0 for remedy) were excluded from further analysis to maintain the integrity of the ground truth labels.

With the ground truths labels established, we now compare against the LLM categorizations. Using Cohen’s Kappa, we observed almost perfect agreement across all three classification tasks, with coefficients of 0.82 for product classification, 0.91 for hazard classification, and 0.90 for remedies classification. These strong agreement levels indicate that the LLM’s predictions align closely with human judgment, achieving a level of consistency comparable to expert annotators. Given these results, we are confident in the robustness and accuracy of the LLM-generated outputs and proceed to use them for subsequent analyses in this paper.

\begin{figure*}[t]
\begin{center}
\includegraphics[width=0.95\linewidth]{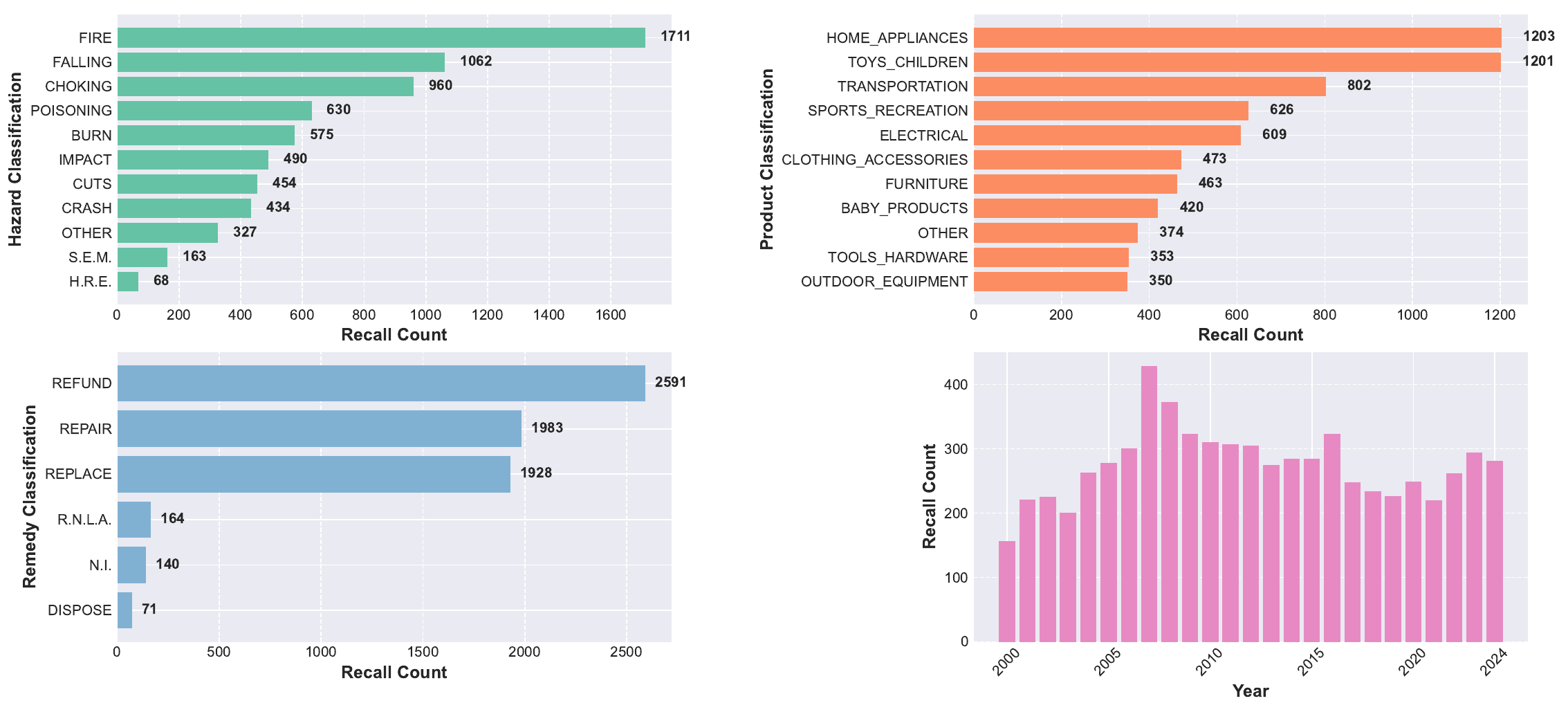}
\caption{Data metrics from 6,874 recalls spanning 2000 - 2024 recall dates.}
\label{fig:fig1}
\end{center}
\end{figure*}

\subsection{Analysis of Recall Classifications}

To analyze patterns in product recalls, we first aggregated the dataset across four dimensions: hazard classifications, product categories, remedy classifications, and recall year. These aggregations are visualized in Fig. ~\ref{fig:fig1}, allowing us to identify prevalent hazards, frequently recalled product types, common industry remedies, and temporal trends in recall activity.

\begin{figure}[t]
\begin{center}
\includegraphics[width=0.55\linewidth]{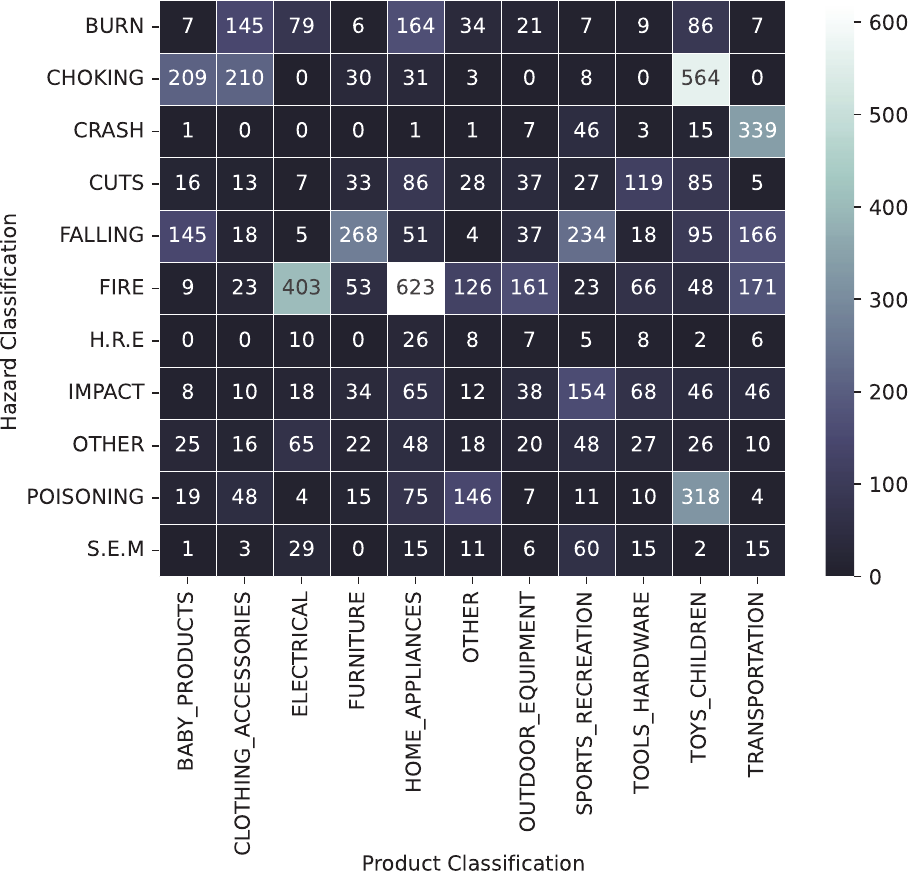}
\caption{Correlation matrix of hazards and product classifications.}
\label{fig:heatmap}
\end{center}
\end{figure}

Additionally, to explore relationships between product categories and associated hazards, we generated a hazard-product co-occurrence heatmap (Fig. ~\ref{fig:heatmap}). This visualization highlights where certain hazards are disproportionately concentrated within specific product types, offering insight into recurring failure modes and industry-specific safety concerns.

Examining \textbf{hazard classification}, \textit{fire}, \textit{falling} and \textit{choking} emerged as the most prevalent hazards, suggesting a need for improved foresight in anticipating these failure modes. Focusing on \textit{fire}, we also see from Fig. \ref{fig:heatmap} that the strongest correlations come from \textit{electrical} and \textit{home\_appliances}. This aligns with existing literature indicating heightened risks of household fires due to electrical failures in sockets, plugs, and wiring, as opposed to householder carelessness \cite{taylor2024electrical}.

Within \textbf{product classification}, the high frequency of recalls in \textit{home\_appliances} and \textit{toys\_children} indicates particular vulnerability to hazards faced within the average US household, indicating the pressing need for safer design of products intended for vulnerable or high-use demographics. A study conducted by Anwar \cite{anwar2014product} also relied on the CPSC database to examine the harms resulting from high recalls in the toy industry. This study found that while most toys were manufactured in China, a vast quantity of toys were designed in the US, leading to harms related to choking and lead poisoning as primary concerns. This analysis aligns with our findings, emphasizing the importance of stronger safety precautions when designing consumer products. Interestingly, the heatmap also shows lower recall frequencies for categories such as \textit{tools\_hardware} and \textit{outdoor\_equipment}, suggesting a possibility of heightened risk awareness and more conservative design practices within these industries.

The prevalence of \textit{refund} as a remedial action indicates a preference towards immediate consumer safety, likely chosen when repairs or replacements are insufficient for risk mitigation. The substantial use of \textit{repair} and \textit{replace} remedies further suggests a widespread industry practice of addressing safety concerns through corrective product interventions rather than solely financial compensation. Kubler et al. \cite{kubler2012impact} conducted a study to observe the effect of product recalls on brand loyalty. Results found that consumers valued transparency and convenient handling of the recalled product.

The temporal analysis revealed an apparent peak in recall incidents around 2005, followed by a general downward trend with fluctuations thereafter. This trend may reflect enhanced regulatory interventions, evolving industry standards, or changes in manufacturing practices and quality control measures. Notably, the recent uptick observed post-2020 signals the potential effects of disruptions in supply chains due to global events.

Taken together, these results emphasize key areas for improved product safety. Recalls in categories such as \textit{home\_appliances}, \textit{electrical}, and \textit{toys\_children} underscore the need for more proactive hazard anticipation during product design. Furthermore, the correlation patterns revealed by the heatmap suggest potential value in cross-domain learning: designers may benefit from examining hazard trends in adjacent product sectors to better anticipate potential risks.

\begin{table}[b]
    \centering
    \renewcommand{\arraystretch}{1.1}
    \caption{Recall description diversity of each product category as measured by Convex Hull area in 2D scaled embedding space.}
    \begin{tabular}{lr}
        \hline
        \textbf{Category} & \textbf{Normalized Area} \\
        \hline
        TOYS\_CHILDREN         & 11.848 \\
        SPORTS\_RECREATION     & 11.009 \\
        TRANSPORTATION        & 10.486 \\
        HOME\_APPLIANCES       & 10.169 \\
        TOOLS\_HARDWARE        & 9.776  \\
        OUTDOOR\_EQUIPMENT     & 9.485  \\
        BABY\_PRODUCTS         & 9.219  \\
        FURNITURE             & 9.111  \\
        OTHER                 & 8.872  \\
        CLOTHING\_ACCESSORIES  & 8.712  \\
        ELECTRICAL            & 6.688  \\
        \hline
    \end{tabular}
    \label{tab:convex_hull_normalized_area}
\end{table}

\subsubsection{Case Study 1: 2D Latent Space of Recall Descriptions}
\begin{figure}[t]
\begin{center}
\includegraphics[width=0.75\linewidth]{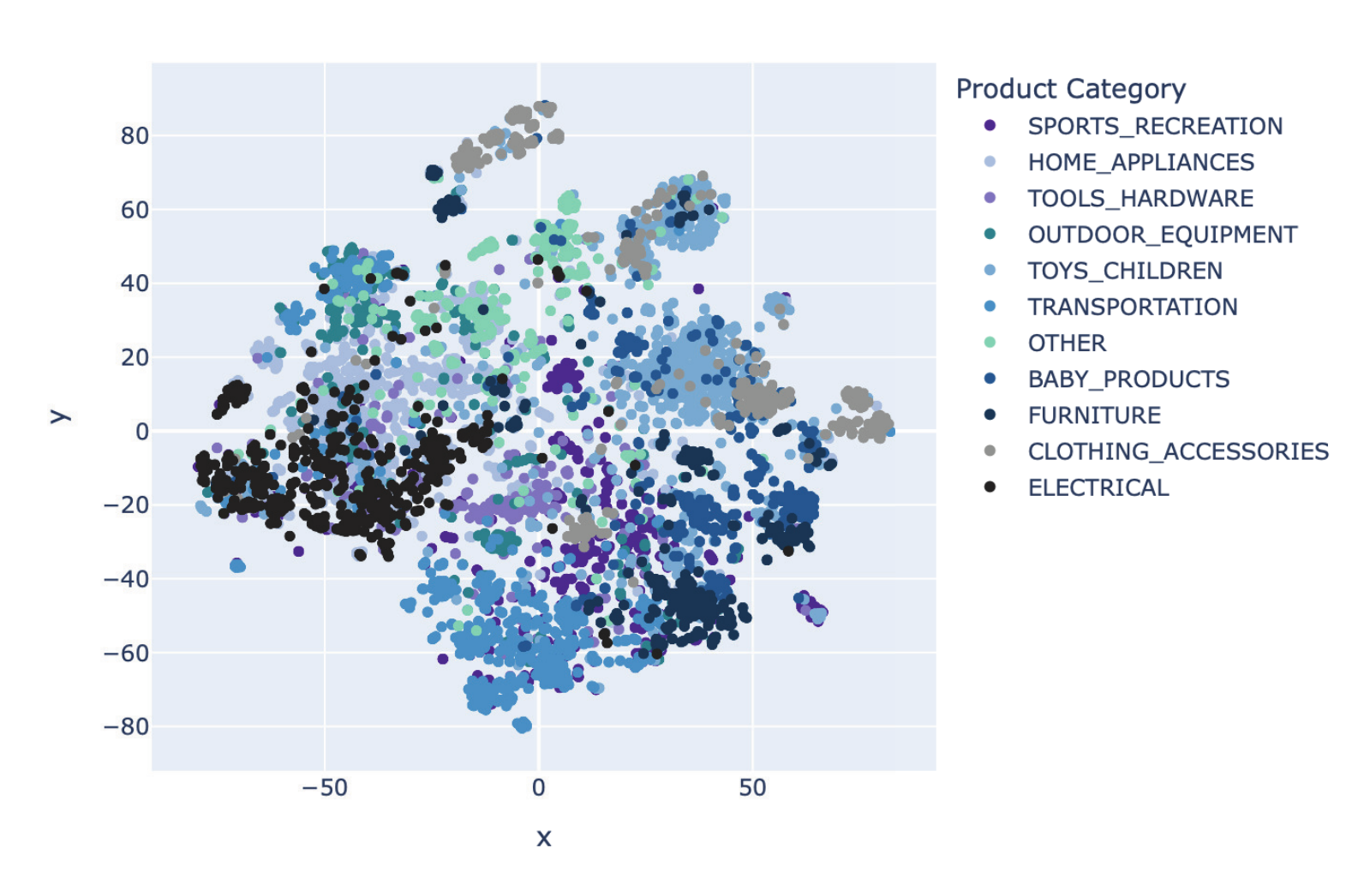}
\caption{Embedding space of recall descriptions labeled by product categories.}
\label{fig:2Dembed}
\end{center}
\end{figure}

In this case study, we examine the embedding space of the \textit{recall\_description} field to explore how textual recall data can reveal underlying patterns beyond predefined hazard classifications. We specifically chose to embed \textit{recall\_description} to augment and challenge existing groupings, offering an opportunity to uncover nuanced relationships within the dataset. Figure~\ref{fig:2Dembed} presents a latent space representation of all 6,874 \textit{recall\_description} texts, revealing natural clusters that broadly correspond to different product categories.

\begin{figure*}[t]
\begin{center}
\includegraphics[width=\linewidth]{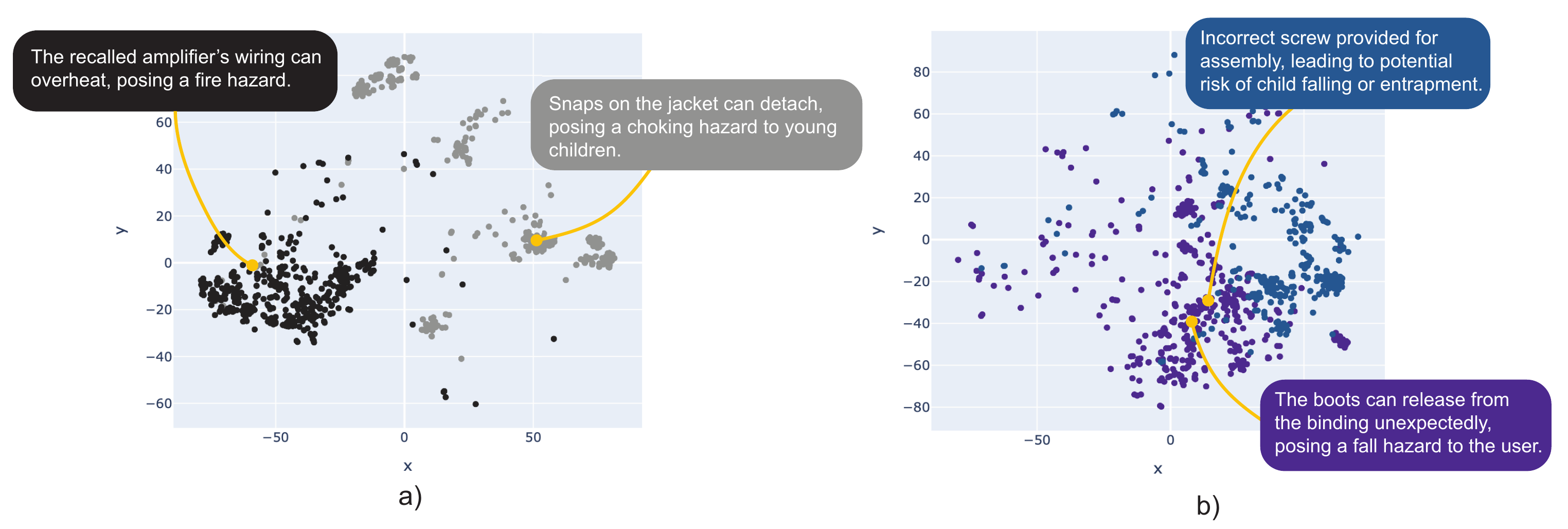}
\caption{a) Embedded recall descriptions of \textit{electrical} (black) and \textit{clothing\_accessories} (gray). b) Embedded recall descriptions of \textit{baby\_products} (blue) and \textit{sports\_recreation} (purple).
Descriptions are denoted for each example, showcasing distant (a) and near (b) recall descriptions across different product categories.}
\label{fig:example_2d}
\end{center}
\end{figure*}

To illustrate how product domains influence the structure of this space, we highlight two specific examples in Fig.~\ref{fig:example_2d}. In Fig.~\ref{fig:example_2d}a, we focus on the product categories \textit{electrical} (black) and \textit{clothing\_accessories} (gray). Here, we observe minimal overlap in the embedding space, reflecting distinct recall descriptions and associated hazards. For example, recalls within the \textit{electrical} category predominantly reference fire hazards, resulting in a more cohesive clustering pattern. In contrast, the \textit{clothing\_accessories} category exhibits multiple distinct clusters, reflecting a wider variety of recall reasons, such as choking hazards due to detachable components. This divergence underscores how certain domains exhibit unique, domain-specific risks, while others exhibit larger diversity in risk possibilities. 

Figure~\ref{fig:example_2d}b highlights a case where recall descriptions from different domains show considerable overlap. Specifically, the categories \textit{baby\_products} and \textit{sports\_recreation} share similarities in recall descriptions, often referencing issues related to unsecured assembly constraints that pose risks of falling or entrapment. Despite the differing nature of these product types, the commonality in risk profiles suggests valuable cross-domain learning opportunities. Designers working in either space could benefit from studying hazards in the other, enabling a more comprehensive approach to safety.

To quantitatively contextualize these findings, we calculate the spread of recall descriptions for each product category using Convex Hull analysis. Table~\ref{tab:convex_hull_normalized_area} reports the normalized areas for each category, offering a metric for the diversity of recall descriptions. This approach, informed by prior work on diversity metrics in embedding spaces \cite{ma2025large}, enables a comparative assessment of risk variability across categories. Notably, the \textit{toys\_children} category exhibits the largest normalized area, indicating a wide range of distinct hazards associated with children's products. This reinforces the need for cautious safety considerations in the design of children's toys.

\begin{figure*}[t]
\begin{center}
\includegraphics[width=\linewidth]{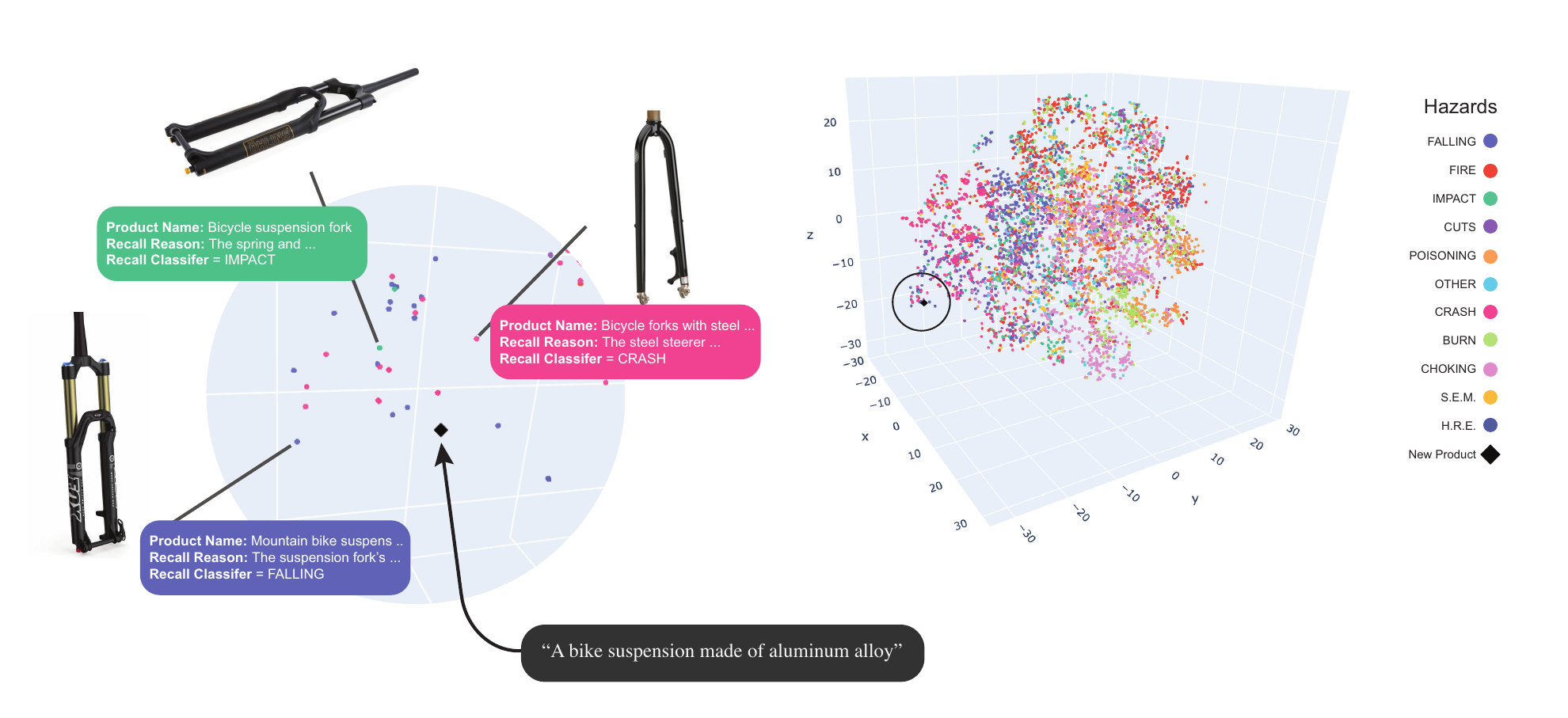}
\caption{3D embedding space of product name colored by hazard class. An example is shown identifying similarities between a new product and existing products with recall information provided.}
\label{fig:3d_embeddings}
\end{center}
\end{figure*}

\subsubsection{Case Study 2: 3D Latent Space of Product Name}
We demonstrate the potential to move beyond passive data exploration by developing an interactive visualization tool that situates new product ideas within the context of historical recall data (see Fig.~\ref{fig:3d_embeddings}). This visualization embeds all product descriptors—specifically \textit{product\_name}—into a shared three-dimensional latent space using t-SNE. When a designer inputs a new product concept, the system embeds the input text into the same space and projects it alongside past recalled products. We chose to embed \textit{product\_name} as it typically conveys precise, yet high-level semantic information, making it particularly suitable for early-stage ideation.

For instance, an engineering team might propose a product idea described simply as \textit{"a bike suspension made of aluminum alloy"} without fully developed specifications. This tool allows them to position that idea within the landscape of similar historical recalls, providing insight into neighboring products and their associated hazard classifications. By visualizing these relationships interactively, designers can proactively identify potential risks, explore relevant precedents, and make more informed design decisions. Ultimately, this integration of recall data into the early design process supports safer, more responsible product development.

\subsubsection{Case Study 3: LLM Hazard Prediction}
In the third case study, we evaluate the ability of LLMs to predict potential hazards based solely on a product's text description of the image. Using the approach detailed in Section 3.3, the LLM analyzes the \textit{visual\_product\_description} field and outputs hazard classifications drawn from the predefined list of categories.

\begin{figure}[t]
\begin{center}
\includegraphics[width=.5\linewidth]{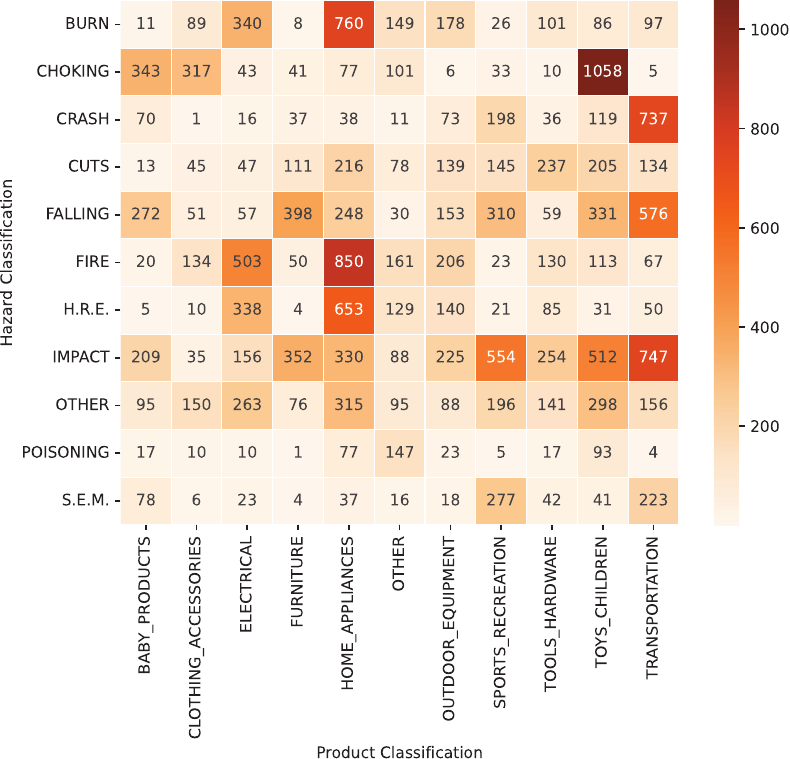}
\caption{Heatmap showing correlations between LLM predicted product and hazard recalls.}
\label{fig:heatmap_llm}
\end{center}
\end{figure}

The aggregated frequencies of the LLM-predicted hazard classes are shown in Fig. \ref{fig:heatmap_llm}. The predicted patterns align closely with actual recall distributions across both product and hazard categories, as seen in Fig. \ref{fig:heatmap}.

To assess predictive performance, we compute the RA metric across all hazard classes. The per-class RA scores are summarized in Table \ref{tab:accuracy_per_class}. The results indicate strong predictive capabilities in several hazard categories, with particularly high RA scores observed for \textit{choking} (0.93) and \textit{crash} (0.91) hazards. Conversely, the poisoning hazard class yields the lowest RA score of 0.32.

\begin{table}[b]
    \centering
    \renewcommand{\arraystretch}{1.1}
    \caption{Accuracy per class and overall relaxed accuracy.}
    \begin{tabular}{l r}
        \hline
        \textbf{Hazard Classification} & \textbf{Relaxed Accuracy} \\
        \hline
        BURN & 0.59 \\
        CHOKING & 0.91 \\
        CRASH & 0.93 \\
        CUTS & 0.65 \\
        FALLING & 0.85 \\
        FIRE & 0.74 \\
        H.R.E. & 0.74 \\
        IMPACT & 0.87 \\
        OTHER & 0.46 \\
        POISONING & 0.32 \\
        S.E.M. & 0.49 \\
        \hline
        \textbf{Overall Relaxed Accuracy} & 0.73 \\
        \hline
    \end{tabular}
    \label{tab:accuracy_per_class}
\end{table}

Further examination of Fig. \ref{fig:heatmap_llm} reveals that the model rarely predicts poisoning hazards for products such as children's toys. However, cross-referencing with actual recall data (Fig. \ref{fig:heatmap}) shows a substantial number of poisoning-related recalls, particularly within the children’s toys category. This discrepancy highlights a critical limitation: certain hazards, like poisoning, may not be visually apparent and thus are underrepresented in LLM predictions. The model’s difficulty in predicting non-visible hazards mirrors how consumers often rely on visual inspection to assess product safety. This highlights a critical need for both transparent hazard communication and proactive design strategies that address risks not readily apparent through appearance alone, particularly in preventing latent hazards such as poisoning.

\section{Dataset Use Cases and Future Applications for Engineering Design}
There are several promising avenues to extend and apply this work. First, augmenting the CPSC dataset with globally recalled product data, such as the OECD Global Recalls Portal \cite{oecd_global_recalls}, would allow for a more comprehensive cross-national analysis. This could reveal broader patterns in product failures and facilitate comparative studies across regulatory environments and cultural contexts.

Additionally, while the methods presented demonstrate feasibility in organizing and classifying recall data, validating their effectiveness in real-world design and safety processes remains an open opportunity. Future studies could engage with industry practitioners to assess how historical recall data—structured and augmented as shown here—can be integrated into existing risk analysis workflows. Specifically, measuring where practitioners derive value would provide actionable insights.

The multimodal nature of the dataset could also support additional work with visual language models (VLMs). VLMs could be used to predict a product's hazard and recall risk, similar to Case Study 3 but with additional visual cues from the product's image. Future studies could determine whether visual or textual cues provide stronger signals for this task.

Beyond retrospective analysis, there is potential to apply this dataset and classification pipeline proactively. For instance, integrating recall-informed hazard classifications into early-stage product requirement generation may improve safety considerations from the outset. This could involve coupling the dataset with LLM generated user requirements \cite{ataei2024elicitron} to ensure risk factors identified in past recalls are embedded into design requirements.

The dataset can further support developing detailed user personas based on specific recall incidents and hazard types (integrating into works such as \cite{ataei2024elicitron}). By examining product recalls through the lens of potential user interactions, designers could construct user personas representative of individuals most likely to encounter or exacerbate certain hazards. For example, examining products recalled due to choking hazards might inform the creation of personas representing families with young children or elderly individuals with limited mobility, allowing designers to conduct roleplay analysis and proactively consider how different user behaviors and demographics might interact with products to induce hazardous situations.

Although it was not used in this work, the \textit{remedy\_classification} metadata included in the dataset could be used to train a model useful to design practitioners in determining an appropriate solution once a recall-level hazard has been found.

Further, future work could explore embedding additional multimodal fields, such as the \textit{visual\_product\_description}, into a unified vector space. Doing so would support more detailed similarity analyses, helping designers quickly identify potential risks based on visual and textual product features. Investigating the downstream implications of these embeddings -- such as their application in automated hazard detection or AI-augmented design tools -- presents another valuable research direction.

Ultimately, by refining these methods and integrating them into decision-making pipelines, this research could contribute to addressing future challenges in product safety, thereby encouraging a practice of proactive, data-driven risk mitigation in product design and development.

\section{Conclusion}
This work demonstrates the feasibility and value of leveraging historical product recall data to identify potential hazards in consumer products. By analyzing recall records, we aim to provide designers and engineers with actionable insights into common failure modes and safety risks, ultimately informing safer and more robust product development. Specifically, we advocate for the integration of publicly available datasets into the early stages of the design process, where risk identification is often most critical yet under-informed.

One of the key challenges in early-stage design is anticipating latent hazards that may not be immediately apparent. Through three distinct case studies, we illustrate the utility of computational and LLM-driven methods for interacting with the dataset. These case studies highlight different modalities of engagement: analyzing textual recall descriptions, embedding product names for similarity assessment, and predicting potential hazards from images. These studies demonstrate approaches for understanding not only \textit{what} products fail, but \textit{how} those failures manifest.

By integrating historical recall data into the design process, we present a scalable and data-driven approach to improve product safety, anticipate failure modes, and support risk-informed decision-making. This research lays the groundwork for future efforts aimed at embedding recall-informed analyses into design workflows, ultimately fostering a culture of proactive, data-supported risk mitigation in engineering design.

\section{Acknowledgement}
Authors DB and KGL would like to acknowledge financial support from the Society of Hellman Fellows and Autodesk Research. Any opinions, findings, and conclusions or recommendations expressed in this article are those of the
authors and do not necessarily reflect the views of the aforementioned sponsors.

\bibliographystyle{unsrt}  
\bibliography{main}  






\appendix       
\clearpage
\section*{Appendix A}
\FloatBarrier  
\begin{table}[h]
    \centering
    \renewcommand{\arraystretch}{1.5}
    \caption{Example of a recall entry from the curated dataset.}
    \begin{tabular}{>{\raggedright\arraybackslash}m{5.5cm} m{10cm}}
        \toprule
        \textbf{Field} & \textbf{Value} \\
        \midrule
        \texttt{recall\_number} & 14259 \\
        \texttt{recall\_date} & 2014-08-20 \\
        \texttt{recall\_description} & The length adjustment buckles release unexpectedly, causing the item being stored to fall and injure people nearby. \\
        \texttt{product\_name} & Kayak and watersports storage hanger \\
        \texttt{product\_quantity} & 10,000 \\
        \texttt{remedies} & Consumers should stop using the recalled storage hangers and return them to the place of purchase for a full refund or replacement. \\
        \texttt{visual\_product\_description} & The product consists of a pair of straps made from blue and black fabric, each approximately 1 inch wide and 84 inches long when unbuckled. They feature plastic snap buckles for length adjustment and plastic-coated steel S-hooks for hanging. \\
        \texttt{product\_classification} & SPORTS\_RECREATION \\
        \texttt{hazard\_classification} & FALLING \\
        \texttt{remedies\_classification} & REPLACE \\
        \texttt{product\_image} & \includegraphics[width=0.2\textwidth]{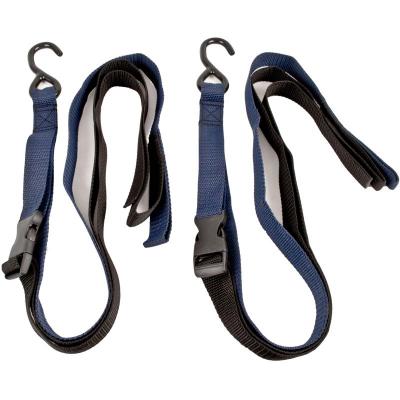} \\ 
        \bottomrule
    \end{tabular}
    \label{tab:recall_data}
\end{table}

\appendix
\clearpage
\section*{Appendix B}
\FloatBarrier 
\begin{table}[h]
    \centering
    \renewcommand{\arraystretch}{1.5}
    \noindent\caption{Hazard classifications and definitions.}
    \begin{tabular}{p{5.5cm} p{10cm}} 
        \hline
        \textbf{Hazard Classification} & \textbf{Definition} \\
        \hline
        Fire & Use of the product may lead to a fire or the product violates federal fabric flammability regulations. \\
        Burn & Use of the product may lead to experiencing burns. \\
        Heat-Related Explosion (H.R.E.) & The product may explode unintentionally. \\
        Falling & Use of the product may cause an unintentional fall. \\
        Poisoning & Use of the product may lead to poisoning. \\
        Crash & Use of the product may lead to an unintentional crash. \\
        Choking & Use of the product may lead to choking, or the product violates federal toy safety standards, or the product violates federal children clothing standards (drawstrings). \\
        Cuts & Use of the product may lead to unintentional cuts and/or lacerations. \\
        Safety Equipment Malfunction (S.E.M.) & The safety product does not operate as intended and use of the product may lead to injury or death. \\
        Impact & Use of the product may lead to an unintentional impact that may cause injury or death. \\
        \hline
    \end{tabular}
    \label{tab:hazard_classifications}
\end{table}

\begin{table}[h]
    \centering
    \renewcommand{\arraystretch}{1.5}
    \noindent\caption{Remedy classifications and definitions.}
    \begin{tabular}{p{5.5cm} p{10cm}} 
        \hline
        \textbf{Remedy Classification} & \textbf{Definition} \\
        \hline
        Refund & A customer may receive a full or partial refund, or gift card for the recalled product. \\
        Repair & The company is offering a repair to the recalled product. \\
        Replace & The company is offering a replacement for the recalled product in the form of a new product or other products of similar value. \\
        Dispose & The product should be thrown out or recycled. \\
        New Instructions (N.I.) & The company will issue new instructions on how the customer can make the recalled product safe. \\
        Remedy No Longer Available (R.N.L.A.) & The recalled product should be thrown out or recycled. \\
        \hline
    \end{tabular}
    \label{tab:remedies_classifications}
\end{table}

\end{document}